%% file: main.tex
\documentclass[letterpaper, 10 pt, conference]{ieeeconf}  

\IEEEoverridecommandlockouts                              

\usepackage{graphics} 
\usepackage{epsfig} 
\usepackage{times} 
\usepackage{amsmath} 
\usepackage{amssymb}  
\usepackage{hyperref}
\usepackage{booktabs}
\usepackage{xcolor}
\usepackage{algorithm}
\usepackage{algpseudocode}
\usepackage{subcaption}
\usepackage{cite}
\usepackage{graphicx}
\usepackage{cuted}    
\usepackage{capt-of}  
\graphicspath{{fig/}}
\usepackage{cuted}
\usepackage{capt-of}
\usepackage{caption}
\title{\LARGE \bf
EasyMimic: A Low-Cost Framework for Robot Imitation Learning from Human Videos
}

\author{Tao Zhang$^{1*}$, Song Xia$^{1*}$, Ye Wang$^{1*\ddagger}$, Qin Jin$^{1\dagger}$
\thanks{$^{*}$Equal contribution, $^{\ddagger}$Project lead, $^{\dagger}$Corresponding author.}
\thanks{$^{1}$AIM3 Lab, Renmin University of China.}
}

\begin{document}

\maketitle
\thispagestyle{empty}
\pagestyle{empty}

\setlength{\stripsep}{0pt}

\input{sec/0_abs}

\input{sec/1_intro}

\input{sec/2_related}

\input{sec/3_method}

\input{sec/4_exp}

\input{sec/5_conlusion}









\bibliographystyle{IEEEtran}
\bibliography{files/main}


\end{document}

%% file: sec/0_abs.tex
\begin{strip}
\vspace{-25pt}
\centering
\includegraphics[width=\textwidth]{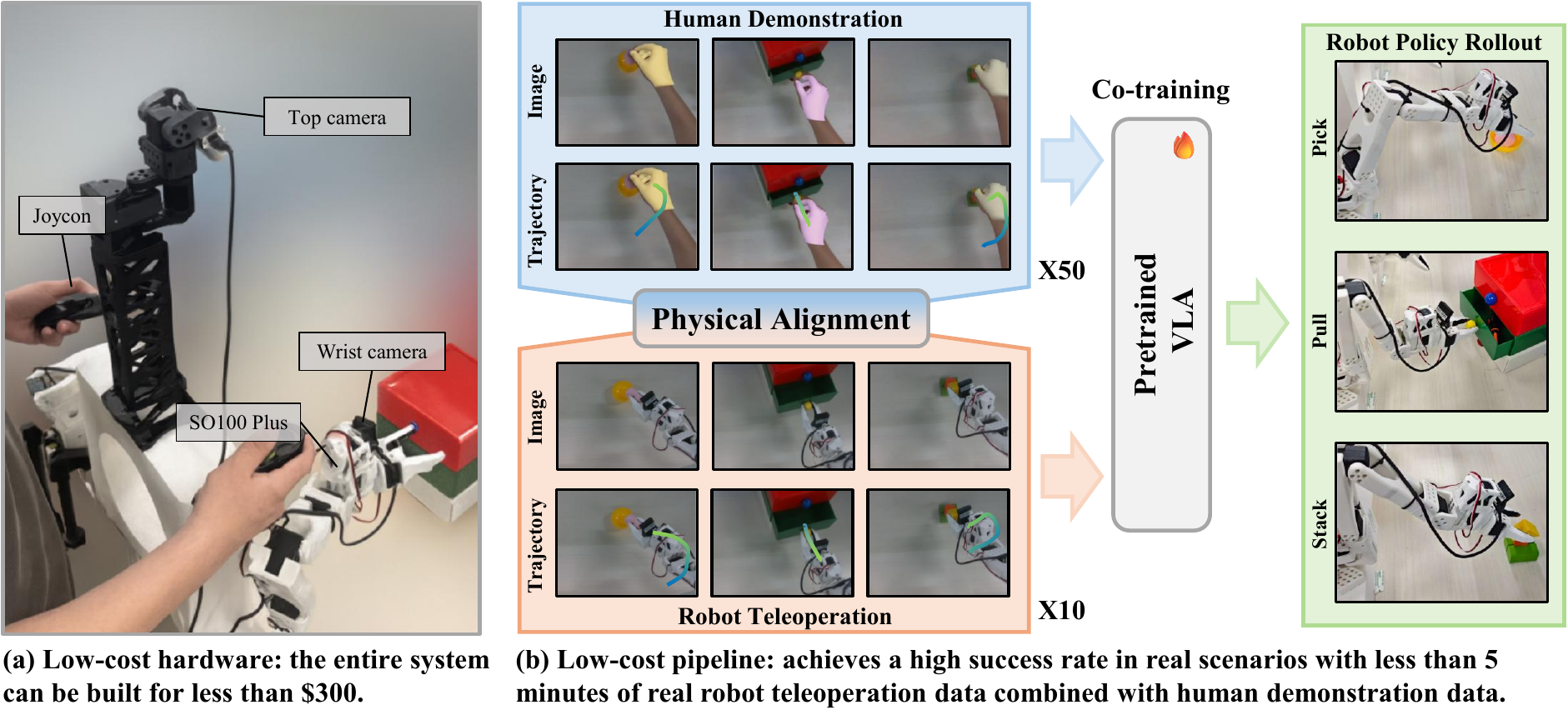}
\captionof{figure}{Overview of the EasyMimic framework. 
The framework learns robotic manipulation from human videos captured with low-cost hardware.
To bridge the embodiment gap, it aligns the action and visual spaces via physical alignment. 
A VLA model is then fine-tuned on the combined data for rapid adaptation to new tasks.}
\label{fig:method_overview}
\end{strip}

\begin{abstract}

Robot imitation learning is often hindered by the high cost of collecting large-scale, real-world data. 
This challenge is especially significant for low-cost robots designed for home use, as they must be both user-friendly and affordable.
To address this, we propose the EasyMimic framework, 
a low-cost and replicable solution that enables robots to quickly learn manipulation policies from human video demonstrations captured with standard RGB cameras.
Our method first extracts 3D hand trajectories from the videos. 
An action alignment module then maps these trajectories to the gripper control space of a low-cost robot. 
To bridge the human-to-robot domain gap, we introduce a simple and user-friendly hand visual augmentation strategy. 
We then use a co-training method, fine-tuning a model on both the processed human data and a small amount of robot data, enabling rapid adaptation to new tasks.
Experiments on the low-cost LeRobot platform demonstrate that EasyMimic achieves high performance across various manipulation tasks. 
It significantly reduces the reliance on expensive robot data collection, offering a practical path for bringing intelligent robots into homes.
Project website: \url{https://zt375356.github.io/EasyMimic-Project/}.

\end{abstract}

%% file: sec/1_intro.tex
\section{INTRODUCTION}

In recent years, bringing robots into homes to assist with daily life has been a long-standing vision in embodied intelligence. 
Equipping robots to master diverse household tasks requires learning from vast amounts of demonstration data. 
Traditional data collection methods, such as expert teleoperation \cite{zhao2023learning,qin2023anyteleop,wu2024gello}, provide high-quality data but are hindered by expensive equipment and complex operation, limiting the widespread adoption of low-cost robots in home environments. 
This data bottleneck remains a core challenges in the field. 
While recent Vision-Language-Action (VLA) models \cite{brohan2023rt1roboticstransformerrealworld,black2024pi0visionlanguageactionflowmodel,kim2024openvla,o2024open,nvidia2025gr00tb} exhibit powerful zero-shot capabilities, unlocking their full potential for precise and reliable manipulation in diverse home settings often requires in-domain fine-tuning. 
This highlights the urgent need for scalable, low-cost data acquisition that can complement these large models.

A promising solution is to leverage ordinary human videos for imitation learning. 
People can easily record manipulation demonstrations using devices like mobile phones, providing a nearly zero-cost and massive-scale data source for robot learning~\cite{kareer2025egomimic,liu2025immimic,qin2022dexmv}. 
Our core motivation is therefore to explore a more convenient and lower-barrier method that allows non-expert users to provide effective demonstration using consumer devices, simplifying and accelerating robot learning.
However, putting this idea into practice requires addressing two fundamental gaps between humans and robots: 
1) \textbf{Visual Appearance Gap}: Human hands differ completely from robot grippers in texture and shape. Models trained directly on human videos tend to overfit these human-specific features, failing to recognize and act correctly when faced with a real robot gripper. 
2) \textbf{Action Space Gap}: The kinematic structures, joint limits, and workspaces of human arms and robot manipulators are vastly different. Directly mimicking human motion trajectories can result in unnatural or even unsafe robot movements. 
While existing work attempts to solve these problems, they often rely on computationally expensive image generation techniques \cite{lepert2025phantom,li2025h2r} or require costly hand-tracking hardware \cite{kareer2025egomimic,qiu2025humanoid,yang2025egovla}, undermining the low-cost and convenience goals for home robots.  

To address the challenges of high data costs and the human-robot embodiment gap, we introduce the \textbf{EasyMimic} framework, a simple and efficient imitation learning framework for low-cost robots using only consumer-grade devices. 
The framework systematically bridges the human-robot gap across two dimensions: action and vision. 
On the action level, we utilize 3D hand pose estimation to extract key kinematic information and design a stable retargeting algorithm to map it into robot actions. 
On the visual level, we discard complex generative models and instead adopt a lightweight visual augmentation strategy. 
By rendering hand meshes with randomized colors, we compel the model to learn cross-embodiment general patterns. 
We then co-train the model on combination of this processed human data and a small amount of robot teleoperation data.
Experiments on a low-cost LeRobot platform demonstrate significant improvements in task success rates.

The main contributions of this work are as follows:
(i) We propose a complete, low-cost pipeline that enables robot policy training from human demonstrations captured with standard RGB cameras.
(ii) We design an action alignment module that effectively retargets 3D human hand trajectories into executable robot actions.
(iii) We employ a lightweight visual augmentation strategy based on hand color randomization to mitigate the visual gap.
(iv) Systematic evaluations on a low-cost robot platform demonstrate that co-training on human data and limited robot data bridges the domain gap and significantly improves task performance.


%% file: sec/2_related.tex
\section{Related Work}

\subsection{Robot Data Collection}
In robotics, teleoperation is a dominant paradigm that offers high safety and low latency, but its data collection efficiency and task coverage are limited \cite{hersch2008dynamical,kormushev2011imitation,li2025train}. 
To improve efficiency and scalability, some methods introduce more advanced data collection hardware or systems \cite{chi2023diffusion,zhao2023learningfinegrainedbimanualmanipulation,ding2024bunny,qin2023anyteleop,chi2024universalmanipulationinterfaceinthewild}. 
However, these approaches typically rely on specialized hardware and skilled operators, incurring significant deployment and maintenance costs.

Hand-centric data collection presents another alternative. 
For example, instrumented gloves can directly record hand poses but suffer from complex calibration and sensor drift \cite{carfi2024modular}. 
Head-mounted devices can capture egocentric views and hand information simultaneously, but their high cost and reliance on SLAM for spatiotemporal alignment limit their broader application \cite{qiu2025humanoid,niu2025human2locoman,liu2025egozero,kareer2025egomimic}.
To address these limitations, our method uses only a single RGB camera to efficiently acquire hand data, striking a better balance between cost and efficiency while achieving high scalability through the rapid iteration of commodity hardware \cite{grauman2024ego}.

\subsection{Learning from Human Videos}
To address the challenges of cross-embodiment learning, prior research can be broadly categorized into three directions. 
The first line of work reduces the appearance gap by synthesizing or editing data to make demonstrations more robot-like, though such pipelines often involve complex rendering and high computational costs \cite{lepert2025phantomb,li2025h2r,christen2024synh2r}. 
The second line focuses on distilling embodiment-agnostic, high-level information from human videos, such as reward functions or sub-task plans, which requires robust object perception and temporal abstraction capabilities \cite{zakka2021xirlcrossembodimentinversereinforcement,wang2023mimicplaylonghorizonimitationlearning,zhu2024visionbasedmanipulationsinglehuman,bharadhwaj2024roboagent}. 
The third line builds unified representations or policies to align human and robot demonstrations at the feature and control levels, thereby mitigating discrepancies in appearance and action jointly\cite{kareer2025egomimic,jain2024vid2robotendtoendvideoconditionedpolicy,zhou2025you,haldar2025point}.
Our work follows this third direction. 
We employ lightweight rendering to standardize hand appearance and narrow the appearance gap, while simultaneously aligning action spaces with inverse kinematics (IK) to map hand trajectories into executable robot actions, enabling low-cost and effective cross-embodiment transfer.

\subsection{Vision Language Action Model}
Driven by the rapid progress of Multimodal Large Language Models (MLLMs), Vision-Language-Action (VLA) models have emerged as a research hotspot in embodied intelligence. By co-training on internet-scale vision-language data and large-scale robot trajectories, these models integrate powerful language understanding and visual perception with robot action generation, achieving unprecedented generalization. Numerous studies have shown that these models exhibit strong potential in open-vocabulary understanding, zero-shot task execution, and long-horizon planning \cite{brohan2023rt1roboticstransformerrealworld,zitkovich2023rt,kim2024openvla,octomodelteam2024octoopensourcegeneralistrobot,nvidia2025gr00tb,black2024pi0visionlanguageactionflowmodel,intelligence2025pi05visionlanguageactionmodelopenworld,peebles2023scalablediffusionmodelstransformers,nair2022r3m,luo2025beingh0}. 
Nevertheless, adapting these general-purpose models to new robot embodiments for specific tasks at low cost remains a critical challenge. 
The core bottleneck lies in the prohibitive expense of collecting demonstration data for each new embodiment. 
Our work directly addresses this challenge by introducing a low-cost, scalable pipeline that leverages human video data as an alternative to expensive robot demonstrations.

%% file: sec/3_method.tex
\section{Method}


We present \textbf{EasyMimic}, a framework that enables low-cost robots to efficiently leverage human video demonstrations.
The overall framework is illustrated in Figure~\ref{fig:method_overview}.
It consists of three core components: 
1) low-cost collection of both human and robot demonstrations using consumer-grade hardware; 
2) physical alignment across the action and visual domains to bridge the embodiment gap; and 
3) a co-training strategy that effectively fuses human and robot data to train a unified policy model.

\subsection{Data Collection Systems and Hardware Design}

In designing the system, we adhere to the principles of low cost and accessibility.
For the robot platform, we use the LeRobot SO100 manipulator \cite{shukor2025smolvlab,wang2025xlerobot}.
To address the limitations of its original five-degree-of-freedom (5-DoF) design in end-effector orientation control, such as singularities in certain workspace areas that prevent arbitrary poses, we added an extra joint, upgrading it to a six-degree-of-freedom (6-DoF) configuration. 
This extension significantly expands its effective workspace.
We name the upgraded manipulator SO100-Plus.
It has a payload capacity of up to 1 kg and a reach of 400 mm, making it well-suited for various tabletop manipulation tasks.

For data collection, we use a Nintendo Joy-Con for robot teleoperation and deploy two consumer-grade RGB cameras: one fixed in a first-person view (shared by human and robot data collection), and the other mounted on the robot's wrist to capture close-up manipulation details.
The entire system can be assembled for under \$300, far more cost-effective than traditional research platforms.
To process human videos, we leverage the advanced HaMeR model \cite{pavlakos2024reconstructing} to extract 3D hand information from monocular inputs. 
We utilize this model to precisely reconstruct the hand morphology for each video frame, providing 21 hand keypoint coordinates $X_t^\mathcal{C}$ and 778 hand mesh vertex coordinates $V_t^\mathcal{C}$ in the camera coordinate frame $\mathcal{C}$.
This information serves as the foundation for our subsequent alignment process, where keypoints are primarily used for action space mapping and the complete mesh is used for visual space alignment.

\subsection{Physical Alignment} 

\begin{figure}[t]
    \centering
    \includegraphics[width=\linewidth]{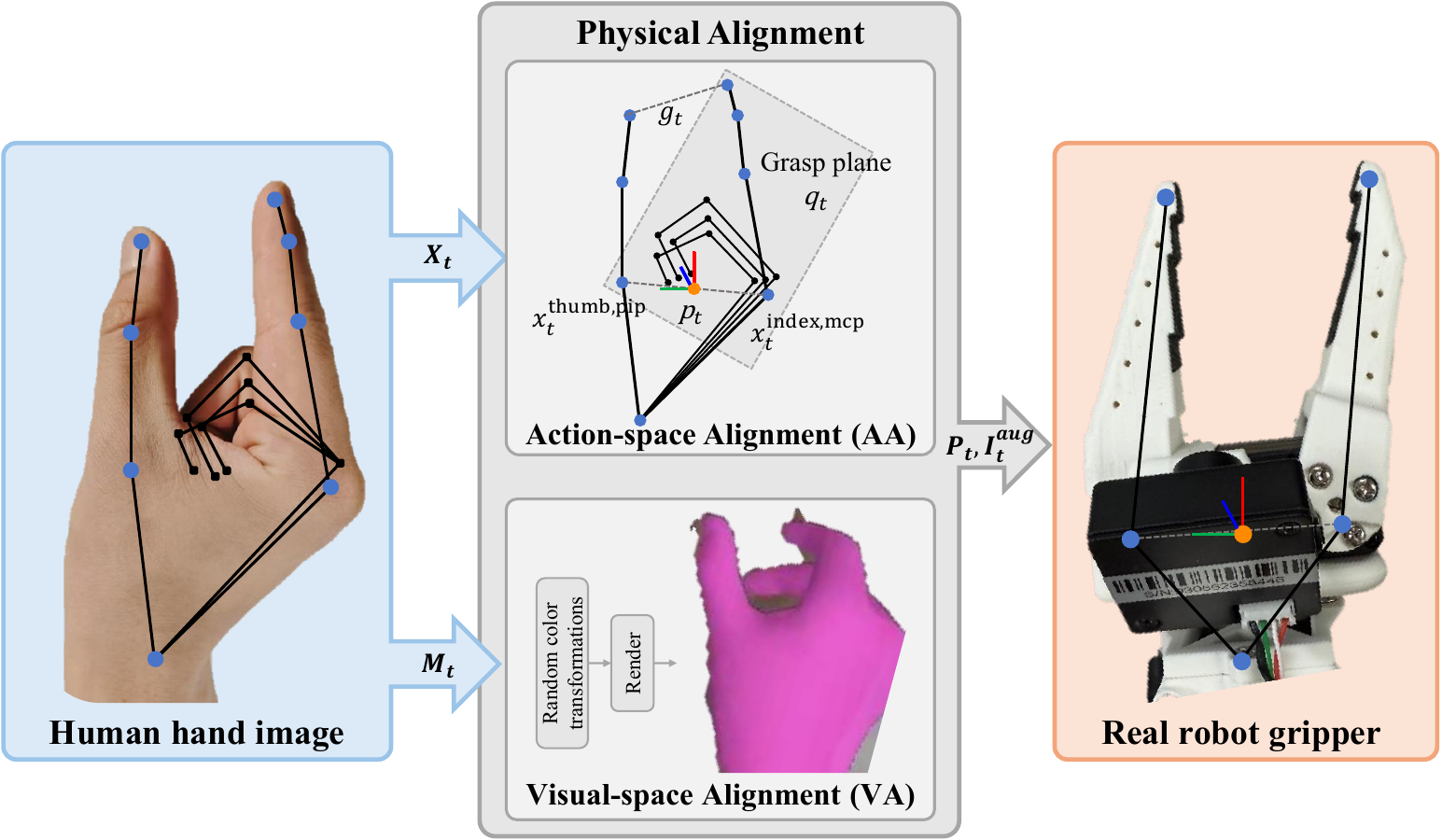}
    \caption{Physical alignment process. 
        Human hand keypoints and meshes are extracted from videos. Hand motion is retargeted to robot actions via the action space alignment module, while the hand mesh is augmented through the visual space alignment module to bridge the physical gap between humans and robots.}  \label{fig:physical_alignment}
\end{figure}

Human-to-robot transfer requires resolving discrepancies in both kinematics and appearance.
As illustrated in Figure~\ref{fig:physical_alignment}, our alignment module systematically addresses these challenges across action space and visual space.

\subsubsection{Action Space Alignment}
The primary challenge in cross-embodiment imitation is to accurately translate human hand motion trajectories into executable action sequences for the robot's end-effector. 
We need to construct a mapping function from human hand kinematics to robot kinematics.
Given a sequence of human hand keypoints $\mathcal{X}_h = \{X_t\}_{t=1}^T$ extracted from a video, our goal is to generate a corresponding sequence of robot end-effector poses $\mathcal{P}_h = \{P_t\}_{t=1}^T = \{(p_t, q_t, g_t)\}_{t=1}^T$, where $p_t$, $q_t$, and $g_t$ represent its 3D position, orientation (typically as a quaternion or Euler angles), and gripper state, respectively.
This pose sequence constitutes the robot's state representation.

\textbf{Position Alignment:} 
Prior works often use the wrist \cite{liu2025immimic} or the midpoint of the fingertips \cite{lepert2025phantomb} as an anchor point, but these can deviate from the true center of interaction during complex manipulations. 
For example, the wrist is too far from the object, while fingertips move relative to each other during fine manipulation, leading to an unstable anchor.
Inspired by unified representation approaches \cite{haldar2025pointb,ren2025motion}, we select the center of the thenar eminence—defined as the midpoint between the thumb's proximal interphalangeal (PIP) joint and the index finger's metacarpophalangeal (MCP) joint—as our retargeting anchor.
This point remains relatively stable during grasping, approximating the palm's center of mass regardless of finger movement. 
It therefore better represents the core region of hand-object interaction.

\textbf{Orientation Alignment:} 
We fit a plane through five keypoints (the four joints of the index finger and the thumb's PIP joint). These five points collectively define the primary orientation of the palm.
The plane's normal vector defines the Z-axis, while the vector from the index finger's MCP to its PIP joint defines the X-axis.
A full 3D coordinate frame is constructed via the cross product, yielding a rotation matrix $R_t$, which is finally converted into the orientation representation $q_t$.

\textbf{Gripper State Alignment:} 
To achieve fine-grained grasping control, we calculate the Euclidean distance $d_t$ between the thumb and index fingertips and normalize it to the range $[0, 1]$ to serve as the gripper state $g_t$:
\begin{equation}
g_t = \text{clip}\left(\frac{d_t - d_{\text{min}}}{d_{\text{max}} - d_{\text{min}}}, 0, 1\right)
\end{equation}
where $d_{\text{min}}$ and $d_{\text{max}}$ are the distance thresholds corresponding to a fully closed and fully open gripper, respectively. 
This continuous gripper state representation, compared to binary open-close control, enables the robot to perform tasks requiring gentle or partial grasps.

After alignment, we use a pre-calibrated hand-eye transformation matrix $T_\mathcal{C}^\mathcal{R}$ to convert the states from the camera frame $\mathcal{P}_h^\mathcal{C}$ to the robot's base frame $\mathcal{P}_h^\mathcal{R}$.
Finally, inspired by chunk-based prediction \cite{zhao2023learningb}, we define the action $a_t$ at timestep $t$ as the state at the next timestep, $P_{h, t+1}$, and form an action chunk $A_t = (a_t, \dots, a_{t+h-1})$ from $h$ future actions. 

\subsubsection{Visual Space Alignment}

The visual discrepancy is another major obstacle in human-to-robot imitation. 
Instead of employing computationally expensive generative models to translate image styles, we propose a lightweight visual augmentation strategy that aligns with the philosophy of our low-cost framework. 
This strategy is based on the idea of domain randomization. Its core objective is to compel the model to ignore task-irrelevant superficial features, such as color and skin texture, and instead learn more fundamental geometric information, such as hand pose and shape.

Specifically, we reuse the extracted 3D hand mesh $M_t$ and render it onto the original image $I_t$. During the rendering process, we apply random color transformations to the entire hand, generating an appearance-standardized augmented view $I_t^{\text{aug}}$. 
By using this augmented data during training, the model is passively exposed to diverse embodiment morphologies. 
This method, without adding any extra complex modules, guides the model at the data level to learn a cross-embodiment visual representation, thereby effectively enhancing its generalization to the robot's morphology. 

\subsection{Training Strategy}

\begin{figure}[t]
    \centering
    \includegraphics[width=\linewidth]{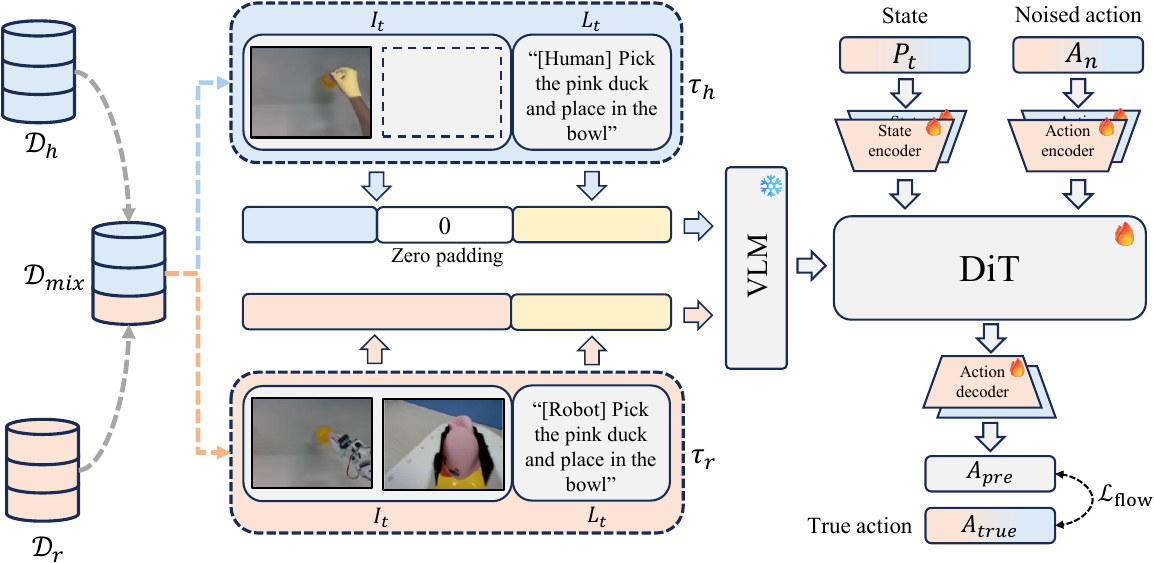}
    \caption{Co-training strategy. Human demonstration data and robot teleoperation data are mixed during training. A shared DiT module learns a unified policy representation, while separate action encoders and decoders for each embodiment handle their specific data properties.  }  
    \label{fig:training_strategy}
\end{figure}

To effectively leverage human and robot data, we adopt a co-training strategy. 
These two data sources have complementary advantages: human data is abundant and easy to acquire, containing rich high-level task semantics, while robot data, though scarce, provides precise low-level control signals consistent with its own kinematic properties. 
We combine our processed human demonstration dataset $\mathcal{D}_h$ with a small real-world robot dataset $\mathcal{D}_r$ to form a mixed dataset $\mathcal{D}_{mix}$. 
As shown in Figure~\ref{fig:training_strategy}, all data first passes through a shared frozen vision-language encoder, followed by a shared Diffusion Transformer (DiT) module for cross-embodiment policy learning.

In our implementation, to handle data imbalance, we use balanced sampling to ensure that each minibatch contains data from both human and robot sources. 
As the human videos lack the wrist-mounted camera view, we apply zero-padding to the corresponding image data to maintain consistent input dimensions.

Furthermore, while the core DiT module is shared, we design independent encoders and decoders for the state ($P_t$) and the noised action ($A_n$) for each embodiment (human and robot). 
This design allows the model to flexibly handle differences in dimensionality, numerical ranges, and physical meanings between the two data types, leading to a more stable shared policy learning process. 

%% file: sec/4_exp.tex
\section{Experiments}

In this section, we first introduce the manipulation tasks and experimental setups. 
We then present a comparison with baseline methods and provide a detailed analysis to validate the effectiveness of our approach.

\subsection{Experimental Setup}
\noindent\textbf{Hardware and Tasks.}
Our experimental platform uses a 6-DoF so100-plus robotic arm equipped with a two-finger gripper.
The vision system includes two monocular RGB cameras: one fixed above the robot's base for a top-down global view, and another mounted on the wrist for an end-effector-centric first-person view.
We evaluate our method on four tabletop manipulation tasks.
\begin{itemize}
    \item \textbf{Pick and Place (Pick):} The robot picks up a toy duck and places it into a bowl. Success is scored in two stages: grasping (0.5) and placing (0.5).
    \item \textbf{Pull and Push (Pull):} The robot must first pull open a drawer and then push it closed. The task is evaluated in two stages: pulling (0.5) and pushing (0.5).
    \item \textbf{Stacking (Stack):} The robot stacks a cube on top of a rectangular block, then stacks a triangular pyramid on top of the cube. Each successful stack is awarded 0.5 points.
    \item \textbf{Language Conditioned (LC):} The robot executes natural language instructions that specify both the target object (e.g., by color) and the placement goal(e.g., target container or location).For example: “pick up the pink duck and place it into the yellow bowl”. Scoring follows the same scheme as the Pick task.
\end{itemize}

\noindent\textbf{Model and Training.}
We use the pre-trained Gr00T N1.5-3B \cite{nvidia2025gr00tb} as the foundation VLA model. 
The policy network is configured to output absolute actions, which consist of 6-DoF end-effector poses and gripper states.
All models are trained for 5,000 gradient steps on a single NVIDIA RTX 4090 GPU using the AdamW optimizer with a learning rate of $1 \times 10^{-4}$ and a batch size of 32.

\noindent\textbf{Data Collection.}
For each task, we collect 100 human video demonstrations and 20 robot teleoperation trajectories.
Table~\ref{tab:data_collection} summarizes the data collection statistics. The human data collection rate is substantially higher than that of robot teleoperation, highlighting the efficiency and utility of leveraging human data.

\begin{table}[htbp]
    \centering 
    \begin{tabular}{c c c c c c c}
        \toprule
        Task & \multicolumn{3}{c}{H} & \multicolumn{3}{c}{R} \\
        \cmidrule(lr){2-4} \cmidrule(lr){5-7}
             & \# & min & \#/min & \# & min & \#/min \\
        \midrule
        Pick & 100 & 8 & 12.5 & 20 & 10 & 2.0 \\
        Pull & 100 & 10 & 10.0 & 20 & 15 & 1.3 \\
        Stack & 100 & 15 & 6.7 & 20 & 40 & 0.5 \\
        LC & 100 & 12 & 8.3 & 20 & 12 & 1.7 \\
        \bottomrule
    \end{tabular}
        \caption{Data collection overview. The table reports the number of demonstrations (\#), total collection time (min), and average collection rate (\#/min) for both human (H) and robot (R) data across different tasks.}  
        \label{tab:data_collection}
\end{table}

\noindent\textbf{Baselines.}
We validate the effectiveness of  EasyMimic, which combines physical alignment with a co-training strategy, by comparing it against the following baselines:
\begin{itemize}
    \item \textbf{Robot-Only (10 traj):} The standard imitation learning baseline, trained solely on 10 robot teleoperation trajectories.
    \item \textbf{Robot-Only (20 traj):} Trained using 20 robot teleoperation trajectories.
    \item \textbf{Pretrain-Finetune:} The model is first pre-trained on processed human video data and then fine-tuned on robot data.
\end{itemize}

\noindent\textbf{Evaluation Metrics.}
We evaluate each task using a stage-based scoring scheme with a maximum score of 1.0. Each task is executed 10 times, and the average score is reported. Performance is assessed with respect to key execution stages, including object localization, grasping, transportation or operation, and final release or reset.

\subsection{Main Results}

\subsubsection{Comparison of Training Strategies}

Table~\ref{tab:training_strategies} compares the performance of different training strategies.
The results show that training with only a small amount of robot data (10 or 20 trajectories) yields limited performance, achieving average success rates of only 0.26 and 0.51, respectively. 
This indicates that even with a pre-trained VLA, training on a small amount of robot data is insufficient to achieve a satisfactory level of task completion. 

In contrast, incorporating human data, either through pre-training or co-training, leads to a substantial performance improvement. Our physical alignment approach combined with the Pretrain-Finetune method achieves an average score of 0.75, significantly outperforming the Robot-Only baselines and demonstrating the value of leveraging human demonstrations. 
However, this two-stage training process is approximately 1.6 times slower than the co-training approach due to its sequential pre-training and fine-tuning phases.
Our \textbf{EasyMimic} framework achieves the best performance across all tasks, with an average score of 0.88, surpassing the Pretrain-Finetune method by 0.13 points and  the Robot-only (10 trajectories) baseline by 0.62 points. 
These results highlight that co-training effectively integrates the strengths of both human and robot data, achieving robust task capabilities with minimal robot data while maintaining efficiency.

\begin{table}[h] 
    \centering
    \setlength{\tabcolsep}{6pt} 
    \renewcommand{\arraystretch}{1.2} 
    \begin{tabular}{lccccc}
        \toprule
        \textbf{Strategy} & \textbf{Pick} & \textbf{Pull} & \textbf{Stack} & \textbf{LC} & \textbf{Avg.} \\
        \midrule
        Robot-Only (10 traj.)         & 0.30 & 0.30 & 0.15 & 0.30 & 0.26 \\
        Robot-Only (20 traj.)         & 0.60 & 0.70 & 0.35 & 0.40 & 0.51 \\
        Pretrain-Finetune             & 0.80 & \textbf{0.90} & 0.50 & 0.80 & 0.75 \\
        \textbf{EasyMimic} & \textbf{1.00} & \textbf{0.90} & \textbf{0.70} & \textbf{0.90} & \textbf{0.88} \\
        \bottomrule
    \end{tabular}
    \caption{Performance evaluation of different training strategies. 
    Results show the average scores across different tasks.}
    \label{tab:training_strategies}
\end{table}

\subsubsection{Language Condition}
The experimental setup of this task includes four target objects: a pink duck, a green duck, a yellow bowl, and a wooden block. 
During data collection, we systematically acquire demonstrations covering four distinct combinations: placing the pink duck into the yellow bowl, placing the pink duck onto the block, placing the green duck into the yellow bowl, and placing the green duck onto the block. 
At inference time, the model executes the specified task following natural language instructions.
As shown in Table~\ref{tab:training_strategies}, our EasyMimic framework significantly improves performance on language-conditioned tasks. Training on 20 robot trajectories alone results in a low score of 0.40. In contrast, EasyMimic, by incorporating human data, boosts the score to 0.90, demonstrating the substantial value of leveraging human demonstrations for complex, language-conditioned manipulation.

\subsection{Further Analysis}

\subsubsection{Effect of Data Scale}
\begin{figure}[h]
    \centering
    \begin{subfigure}[b]{0.5\linewidth}
        \includegraphics[width=\linewidth]{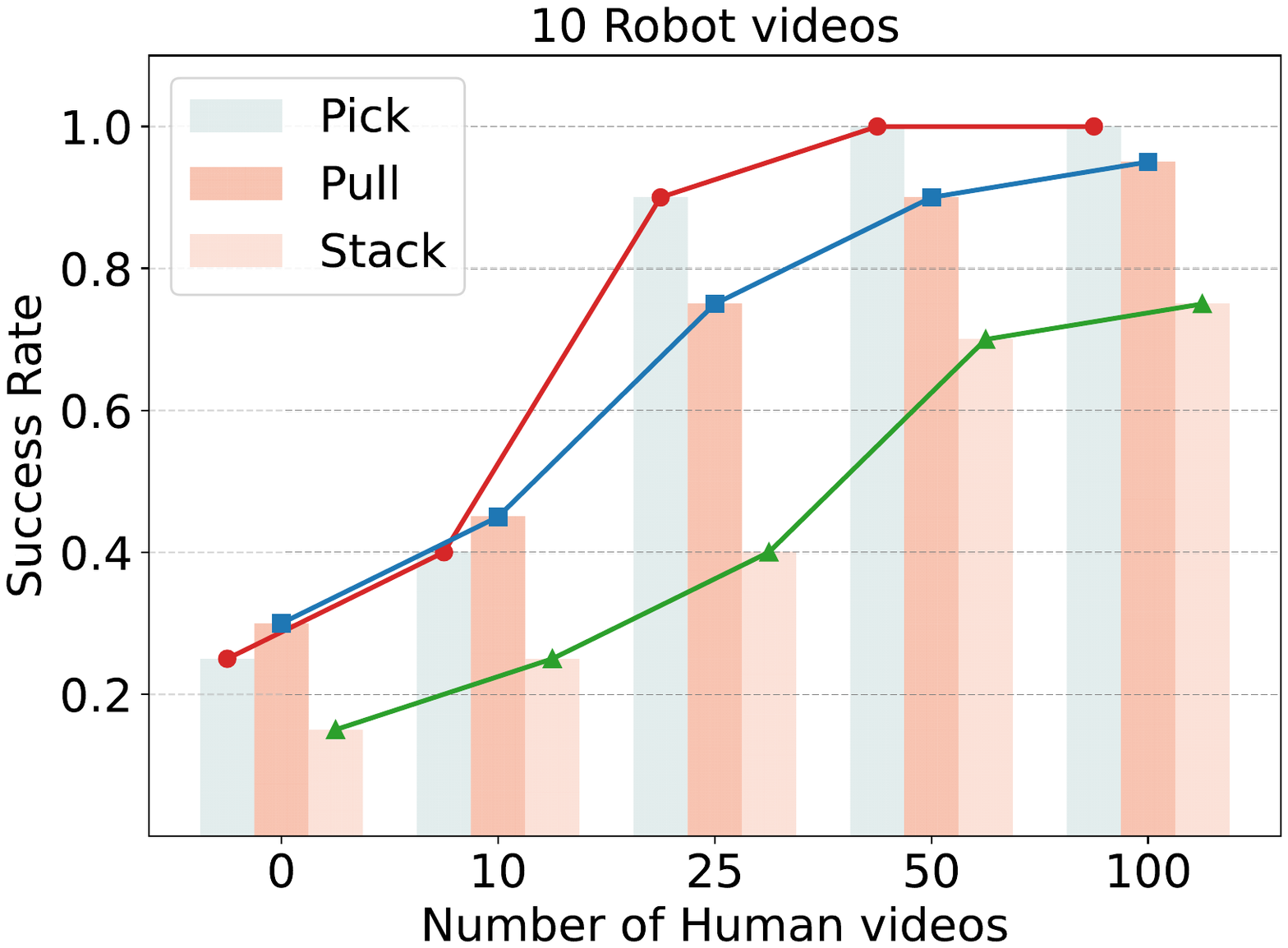}
        \caption{Varying human data}
        \label{fig:human_dataset_size}
    \end{subfigure}
    \hfill
    \begin{subfigure}[b]{0.48\linewidth}
        \includegraphics[width=\linewidth]{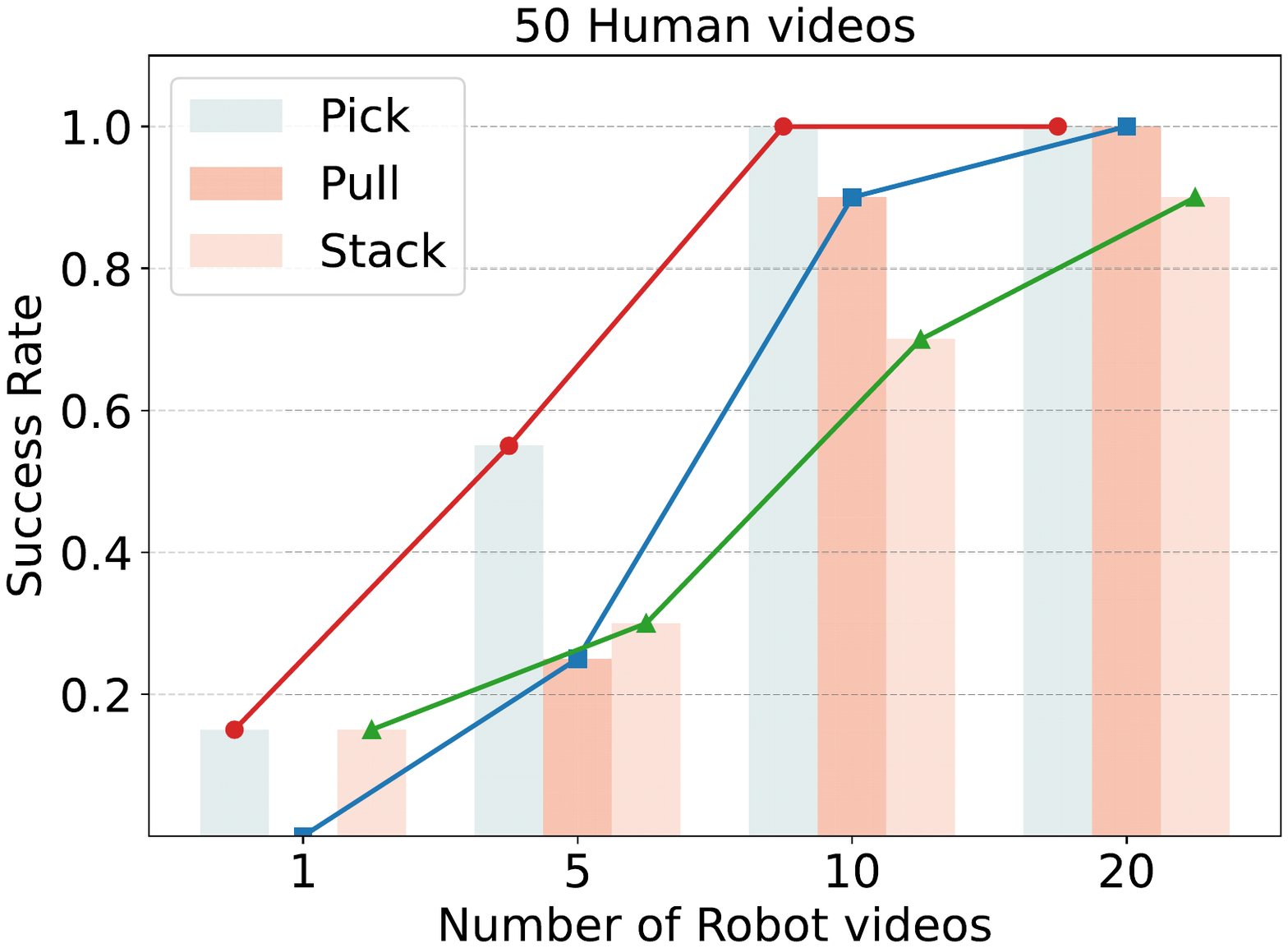}
        \caption{Varying robot data}
        \label{fig:robot_dataset_size}
    \end{subfigure}
    \caption{Effect of Dataset Size. (a) Varying human data with fixed robot data (10 trajectories). 
    (b) Varying robot data with fixed human data (50 videos).} 
    \label{fig:dataset_size}
\end{figure}

Figure~\ref{fig:dataset_size} analyzes the scaling effects of both human and robot data on model performance.

As shown in Figure~\ref{fig:human_dataset_size} increasing the amount of human demonstrations consistently improves performance across all tasks. 
However, we observe diminishing returns beyond 50 demonstrations, suggesting that while human data provides rich and diverse task priors, there exists an optimal scale beyond which additional data yields limited gains.

Figure~\ref{fig:robot_dataset_size} shows the effect of scaling robot data. Performance increases notably when the number of trajectories grows from 5 to 10 but quickly saturates thereafter, indicating that only a small amount of robot data is needed for effective domain adaptation when complemented with abundant human demonstrations.

The experiments confirm that with sufficient human data (e.g., 50 demonstrations), high performance can be achieved using only a small amount of robot data (e.g., 10-20 trajectories), significantly reducing reliance on expensive robot data collection while maintaining robust task execution.

\begin{figure*}[h]
    \centering
    \includegraphics[width=\linewidth]{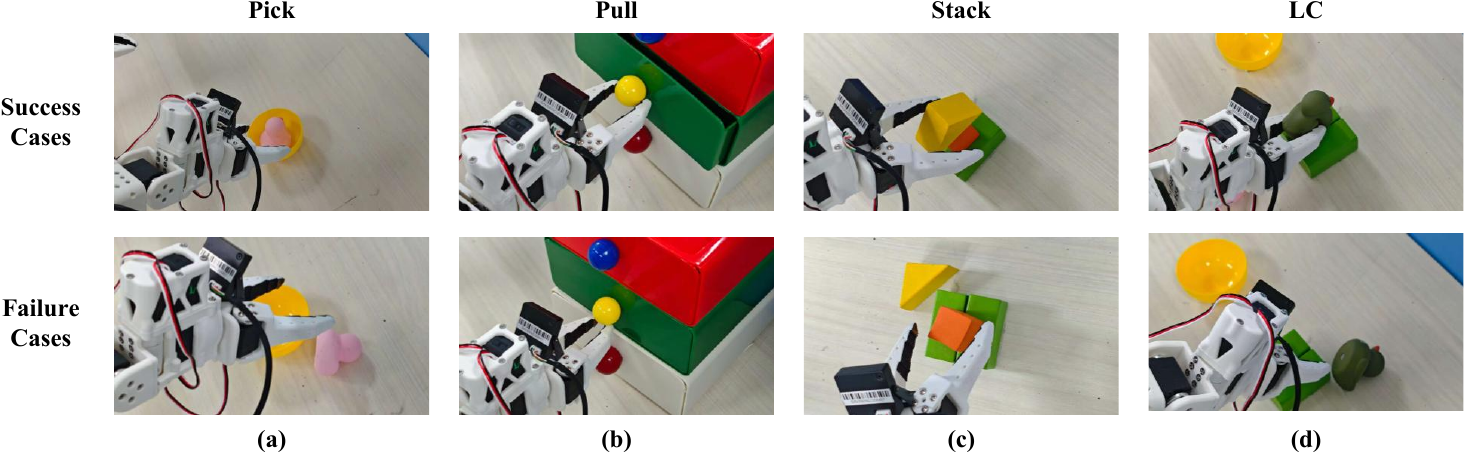}
        \caption{Case analysis of failure modes across different tasks. (a) Premature gripper release during pick and place. (b) Imprecise handle grasping in drawer manipulation. (c) Collision-induced object falling during stacking. (d) Unstable placement leading to the object falling.}
            \label{fig:case_study}
\end{figure*}

\subsubsection{Effect of Alignment}

\begin{figure}[h]
    \centering
    \includegraphics[width=1\linewidth]{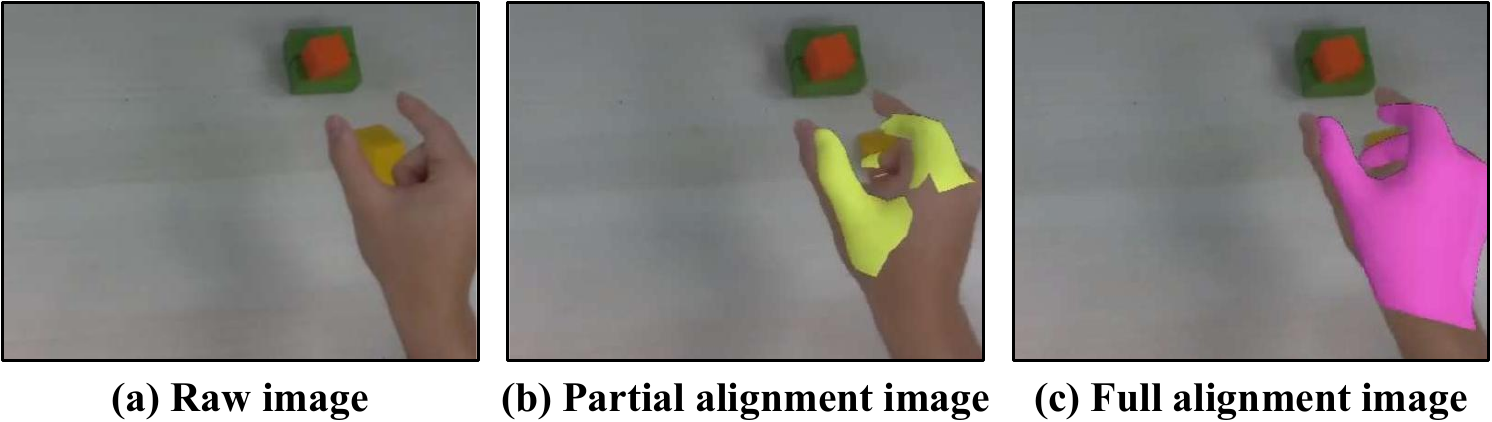}
    \caption{Visualization of the VA module. (a) Original human hand image. (b) Partial masking of the thumb and index finger (VA-Partial). (c) Full masking of the hand region (VA-Full).}
    \label{fig:VA_module}
\end{figure}

Action Alignment (AA) is a core component for translating human motions into executable robot actions.
To quantify its impact, we compare our full EasyMimic pipeline against a baseline where human data is used without proper AA.
As shown in Table~\ref{tab:alignment_ablation}, removing AA leads to a significant performance drop, with an average decrease of 0.27 points across tasks.
The performance degradation is particularly pronounced in tasks such as pick-and-stack, which require precise hand rotation. 
This demonstrates that AA effectively reduces discrepancies between human and robot data, enabling the model to learn more generalizable task representations and strategies.

Visual Alignment (VA) addresses the visual domain gap between human and robot demonstrations. When VA is removed, performance drops sharply, with the average score decreasing by 0.47 points.
This highlights that without visual augmentation, cross-embodiment knowledge transfer is severely hindered. Our lightweight VA module enhances the model’s understanding of shared task context, improving execution capabilities.

We further evaluate the design choice of full versus partial hand masking in the VA module. Figure~\ref{fig:VA_module}(b) illustrates the partial masking strategy (VA-Partial), which only covers the thumb and index finger. As shown in Table~\ref{tab:VA-alignment_ablation}, VA-Partial performs poorly (average score 0.27), indicating that full-hand visual augmentation is crucial for bridging the visual gap and achieving robust cross-embodiment transfer.

\begin{table}[h]
    \centering 
    \begin{tabular}{l c c c c}
        \toprule
        \textbf{Strategy} & \textbf{Pick} & \textbf{Pull} & \textbf{Stack} & \textbf{Avg.}\\
        \midrule
        \textbf{EasyMimic} & \textbf{1.00}  & \textbf{0.90} & \textbf{0.70} & \textbf{0.87}\\
        EasyMimic-AA & 0.50  &  0.80 & 0.50 & 0.60 \\
        EasyMimic-VA &  0.40 &  0.40 & 0.40 & 0.40 \\
        \bottomrule
    \end{tabular}
    \caption{Ablation study on the effectiveness of Action Alignment and Visual Alignment strategies.}
    \label{tab:alignment_ablation}
\end{table}

\begin{table}[h]
    \centering 
    \begin{tabular}{l c c c c}
        \toprule
        \textbf{Strategy} & \textbf{Pick} & \textbf{Pull} & \textbf{Stack} & \textbf{Avg.} \\
        \midrule
        \textbf{EasyMimic} & \textbf{1.00}  & \textbf{0.90} & \textbf{0.70} & \textbf{0.87} \\
        EasyMimic (VA-Partial) & 0.30  & 0.20  & 0.30 & 0.27 \\
        \bottomrule
    \end{tabular}
    \caption{Ablation study on the effectiveness of Visual Alignment strategies.}
    \label{tab:VA-alignment_ablation}
\end{table}

\subsubsection{Effect of Independent Action Heads}
Our EasyMimic framework employs independent action heads for human and robot data to account for their distinct characteristics. To validate this design, we compare it against a variant that employs a single shared action head. As shown in Table~\ref{tab:action_head_ablation}, using independent heads improves the average performance by 0.40 points over the shared-head variant.
This improvement suggests that a shared action head can confuse the model, as it cannot effectively differentiate between the two types of action data. In contrast, explicitly separating the heads prevents interference between human and robot data, allowing the model to leverage the unique properties of each source. This design enhances task execution capabilities by ensuring that both data types contribute effectively to learning a unified policy.

\begin{table}[h]
    \centering 
    \begin{tabular}{l c c c c}
        \toprule
        \textbf{Strategy} & \textbf{Pick} & \textbf{Pull} & \textbf{Stack} & \textbf{Avg.} \\
        \midrule
        \textbf{EasyMimic} & \textbf{1.00}  & \textbf{0.90} & \textbf{0.70} & \textbf{0.87} \\
        EasyMimic (Shared) & 0.60 & 0.40 & 0.40 & 0.47 \\
        \bottomrule
    \end{tabular}
    \caption{Ablation study on the effect of independent versus shared action heads.}
    \label{tab:action_head_ablation}
\end{table}

\subsubsection{Comparison with and without Pretraining}
Both pretraining and incorporating human demonstration data contribute to performance improvement, and their combination proves particularly effective. We evaluate the effect of completely randomly initializing the action expert of GR00T-N1.5-3B. 
As shown in Table~\ref{tab:pretrain_ablation}, training on robot data alone without pretraining yields an average score of only 0.15. Using a pre-trained VLA improves this to 0.25, demonstrating the benefit of large-scale pretraining. Even without pretraining, applying the EasyMimic framework raises the score from 0.15 to 0.53, indicating that EasyMimic is effective even for randomly initialized models. 
Combining EasyMimic with the pre-trained action expert achieves the highest score of 0.87, highlighting the complementary strengths of human demonstration data and pretraining. Notably, relying solely on pretraining without leveraging human videos remains substantially less effective, underscoring the critical role of human data for downstream task adaptation.

\begin{table}[h]
    \centering 
    \begin{tabular}{l c c c c}
        \toprule
        \textbf{Initialization} & \textbf{Pick} & \textbf{Pull} & \textbf{Stack} & \textbf{Avg.} \\
        \midrule
        w/o Pretraining (Robot-Only) & 0.20 & 0.10 & 0.10 & 0.15 \\
        w/o Pretraining (EasyMimic) & 0.70 & 0.50 & 0.40 & 0.53 \\
        w/ Pretraining (Robot-Only) & 0.30 & 0.30 & 0.15 & 0.25 \\
        \textbf{w/ Pretraining (EasyMimic)} & \textbf{1.00} & \textbf{0.90} & \textbf{0.70} & \textbf{0.87} \\
        \bottomrule
    \end{tabular}
    \caption{Effect of pre-trained VLA initialization.}
    \label{tab:pretrain_ablation}
\end{table}

\subsubsection{Generalization to Unseen Objects}
We evaluate zero-shot generalization on the Pick and Place task by training the model exclusively on a pink duck and testing it on two unseen objects: a green duck (unseen color) and a pink cube (unseen geometry and affordance). As shown in Table~\ref{tab:generalization}, EasyMimic outperforms the Robot-Only baseline on both unseen objects, demonstrating effective zero-shot transfer across variations in color and shape. This indicates that leveraging human demonstration data enables the model to learn more generalizable task strategies beyond the objects seen during training.

\begin{table}[h]
    \centering
    \begin{tabular}{lccc}
        \toprule
        \textbf{Strategy} & \textbf{Green duck} & \textbf{Pink cube} & \textbf{Avg.} \\
        \midrule
        Robot-Only & 0.50 & 0.20 & 0.35 \\
        \textbf{EasyMimic} & \textbf{0.80} & \textbf{0.50} & \textbf{0.65} \\
        \bottomrule
    \end{tabular}
    \caption{Generalization performance on unseen objects for the Pick and Place task.}
    \label{tab:generalization}
\end{table}

\subsubsection{Case Analysis}

To gain deeper insight into the model's behavior, we analyze both successful executions and common failure cases across different tasks.
Figure~\ref{fig:case_study} illustrates representative examples of common failure modes: In pick and place tasks, the robot occasionally releases the gripper before reaching the target location, often due to an insufficient understanding of spatial relationships. 
For drawer manipulation, the model sometimes fails to accurately grasp the handle, particularly when the handle's orientation differs from training examples. 
During stacking operations, the robot may inadvertently knock over existing objects while attempting to place new ones, indicating limitations in spatial awareness. 
The model occasionally places objects at unstable positions near edges, leading to subsequent falls and task failures. 
After training with the EasyMimic framework, the model can effectively learn these action logics from human demonstrations, leading to improved task execution and higher success rates.

%% file: sec/5_conlusion.tex
\section{CONCLUSIONS}
In this paper, we introduce the EasyMimic framework, a low-cost, efficient, and replicable paradigm for  robot learning in non-standardized settings. EasyMimic enables robots to acquire manipulation skills by leveraging readily available human videos. 
By incorporating carefully designed action and visual alignment modules, the framework effectively bridges the morphological and kinematic gaps between human demonstrations and robot execution.

Our experimental results demonstrate that co-training on easily accessible human video data, combined with a minimal amount of robot teleoperation data, allows EasyMimic to achieve superior performance across multiple tabletop manipulation tasks. It significantly outperforms baseline methods that rely solely on limited robot data, validating the effectiveness of our approach. This framework provides a practical solution to the data bottleneck problem in robot learning and has potential to lower the barrier for ordinary users to teach robots new skills.


%% file: files/main.bib
@misc{wang2025xlerobot,
    author = {Wang, Gaotian and Lu, Zhuoyi},
    title = {XLeRobot: A Practical Low-cost Household Dual-Arm Mobile Robot Design for General Manipulation},
    howpublished = "\url{https://github.com/Vector-Wangel/XLeRobot}",
    year = {2025}
}

@inproceedings{wu2024gello,
  title={Gello: A general, low-cost, and intuitive teleoperation framework for robot manipulators},
  author={Wu, Philipp and Shentu, Yide and Yi, Zhongke and Lin, Xingyu and Abbeel, Pieter},
  booktitle={2024 IEEE/RSJ International Conference on Intelligent Robots and Systems (IROS)},
  pages={12156--12163},
  year={2024},
  organization={IEEE}
}

@article{lepert2025phantom,
  title={Phantom: Training robots without robots using only human videos},
  author={Lepert, Marion and Fang, Jiaying and Bohg, Jeannette},
  journal={arXiv preprint arXiv:2503.00779},
  year={2025}
}

@misc{haldar2025pointb,
  title = {Point Policy: Unifying Observations and Actions with Key Points for Robot Manipulation},
  author = {Haldar, Siddhant and Pinto, Lerrel},
  year = {2025},
  number = {arXiv:2502.20391},
  eprint = {2502.20391},
  primaryclass = {cs},
  publisher = {arXiv},
  doi = {10.48550/arXiv.2502.20391},
  urldate = {2025-07-01}
}

@inproceedings{kareer2025egomimic,
  title={Egomimic: Scaling imitation learning via egocentric video},
  author={Kareer, Simar and Patel, Dhruv and Punamiya, Ryan and Mathur, Pranay and Cheng, Shuo and Wang, Chen and Hoffman, Judy and Xu, Danfei},
  booktitle={2025 IEEE International Conference on Robotics and Automation (ICRA)},
  pages={13226--13233},
  year={2025},
  organization={IEEE}
}

@misc{lepert2025phantomb,
  title = {Phantom: Training Robots Without Robots Using Only Human Videos},
  author = {Lepert, Marion and Fang, Jiaying and Bohg, Jeannette},
  year = {2025},
  number = {arXiv:2503.00779},
  eprint = {2503.00779},
  primaryclass = {cs},
  publisher = {arXiv},
  doi = {10.48550/arXiv.2503.00779},
  urldate = {2025-07-01}
}

@article{liu2025immimic,
  title={Immimic: Cross-domain imitation from human videos via mapping and interpolation},
  author={Liu, Yangcen and Shin, Woo Chul and Han, Yunhai and Chen, Zhenyang and Ravichandar, Harish and Xu, Danfei},
  journal={arXiv preprint arXiv:2509.10952},
  year={2025}
}

@article{haldar2025point,
  title={Point policy: Unifying observations and actions with key points for robot manipulation},
  author={Haldar, Siddhant and Pinto, Lerrel},
  journal={arXiv preprint arXiv:2502.20391},
  year={2025}
}

@misc{luo2025beingh0,
  title = {Being-H0: Vision-Language-Action Pretraining from Large-Scale Human Videos},
  author = {Luo, Hao and Feng, Yicheng and Zhang, Wanpeng and Zheng, Sipeng and Wang, Ye and Yuan, Haoqi and Liu, Jiazheng and Xu, Chaoyi and Jin, Qin and Lu, Zongqing},
  year = {2025},
  number = {arXiv:2507.15597},
  eprint = {2507.15597},
  primaryclass = {cs},
  publisher = {arXiv},
  doi = {10.48550/arXiv.2507.15597},
  urldate = {2025-07-30}
}

@article{nvidia2025gr00tb,
  title   = {GR00T N1: An Open Foundation Model for Generalist Humanoid Robots},
  author  = {NVIDIA and Bjorck, Johan and Castaneda, Fernando and Cherniadev, Nikita and Da, Xingye and Ding, Runyu and Fan, Linxi Jim and Fang, Yu and Fox, Dieter and Hu, Fengyuan and Huang, Spencer and Jang, Joel and Jiang, Zhenyu and Kautz, Jan and Kundalia, Kaushil and Lao, Lawrence and Li, Zhiqi and Lin, Zongyu and Lin, Kevin and Liu, Guilin and Llontop, Edith and Magne, Loic and Mandlekar, Ajay and Narayan, Avnish and Nasiriany, Soroush and Reed, Scott and Tan, You Liang and Wang, Guanzhi and Wang, Zu and Wang, Jing and Wang, Qi and Xiang, Jiannan and Xie, Yuqi and Xu, Yinzhen and Xu, Zhenjia and Ye, Seonghyeon and Yu, Zhiding and Zhang, Ao and Zhang, Hao and Zhao, Yizhou and Zheng, Ruijie and Zhu, Yuke},
  journal = {arXiv preprint arXiv:2503.14734},
  year    = {2025},
  doi     = {10.48550/arXiv.2503.14734},
  url     = {https://arxiv.org/abs/2503.14734}
}

@misc{ren2025motion,
  title = {Motion Tracks: A Unified Representation for Human-Robot Transfer in Few-Shot Imitation Learning},
  author = {Ren, Juntao and Sundaresan, Priya and Sadigh, Dorsa and Choudhury, Sanjiban and Bohg, Jeannette},
  year = {2025},
  number = {arXiv:2501.06994},
  eprint = {2501.06994},
  primaryclass = {cs},
  publisher = {arXiv},
  doi = {10.48550/arXiv.2501.06994},
  urldate = {2025-07-30}
}

@misc{shukor2025smolvlab,
  title = {SmolVLA: A Vision-Language-Action Model for Affordable and Efficient Robotics},
  author = {Shukor, Mustafa and Aubakirova, Dana and Capuano, Francesco and Kooijmans, Pepijn and Palma, Steven and Zouitine, Adil and Aractingi, Michel and Pascal, Caroline and Russi, Martino and Marafioti, Andres and Alibert, Simon and Cord, Matthieu and Wolf, Thomas and Cadene, Remi},
  year = {2025},
  number = {arXiv:2506.01844},
  eprint = {2506.01844},
  primaryclass = {cs},
  publisher = {arXiv},
  doi = {10.48550/arXiv.2506.01844},
  urldate = {2025-07-01}
}

@misc{zhao2023learningb,
  title = {Learning Fine-Grained Bimanual Manipulation with Low-Cost Hardware},
  author = {Zhao, Tony Z. and Kumar, Vikash and Levine, Sergey and Finn, Chelsea},
  year = {2023},
  number = {arXiv:2304.13705},
  eprint = {2304.13705},
  primaryclass = {cs},
  publisher = {arXiv},
  doi = {10.48550/arXiv.2304.13705},
  urldate = {2025-07-01}
}

@article{hersch2008dynamical,
  title={Dynamical system modulation for robot learning via kinesthetic demonstrations},
  author={Hersch, Micha and Guenter, Florent and Calinon, Sylvain and Billard, Aude},
  journal={IEEE Transactions on Robotics},
  volume={24},
  number={6},
  pages={1463--1467},
  year={2008},
  publisher={IEEE}
}

@article{kormushev2011imitation,
  title={Imitation learning of positional and force skills demonstrated via kinesthetic teaching and haptic input},
  author={Kormushev, Petar and Calinon, Sylvain and Caldwell, Darwin G},
  journal={Advanced Robotics},
  volume={25},
  number={5},
  pages={581--603},
  year={2011},
  publisher={Taylor \& Francis}
}

@article{li2025train,
  title={How to Train Your Robots? The Impact of Demonstration Modality on Imitation Learning},
  author={Li, Haozhuo and Cui, Yuchen and Sadigh, Dorsa},
  journal={arXiv preprint arXiv:2503.07017},
  year={2025}
}

@article{chi2023diffusion,
  title={Diffusion policy: Visuomotor policy learning via action diffusion},
  author={Chi, Cheng and Xu, Zhenjia and Feng, Siyuan and Cousineau, Eric and Du, Yilun and Burchfiel, Benjamin and Tedrake, Russ and Song, Shuran},
  journal={The International Journal of Robotics Research},
  pages={02783649241273668},
  year={2023},
  publisher={SAGE Publications Sage UK: London, England}
}

@misc{zhao2023learningfinegrainedbimanualmanipulation,
      title={Learning Fine-Grained Bimanual Manipulation with Low-Cost Hardware}, 
      author={Tony Z. Zhao and Vikash Kumar and Sergey Levine and Chelsea Finn},
      year={2023},
      eprint={2304.13705},
      archivePrefix={arXiv},
      primaryClass={cs.RO},
      url={https://arxiv.org/abs/2304.13705}, 
}

@article{ding2024bunny,
  title={Bunny-visionpro: Real-time bimanual dexterous teleoperation for imitation learning},
  author={Ding, Runyu and Qin, Yuzhe and Zhu, Jiyue and Jia, Chengzhe and Yang, Shiqi and Yang, Ruihan and Qi, Xiaojuan and Wang, Xiaolong},
  journal={arXiv preprint arXiv:2407.03162},
  year={2024}
}

@article{qin2023anyteleop,
  title={Anyteleop: A general vision-based dexterous robot arm-hand teleoperation system},
  author={Qin, Yuzhe and Yang, Wei and Huang, Binghao and Van Wyk, Karl and Su, Hao and Wang, Xiaolong and Chao, Yu-Wei and Fox, Dieter},
  journal={arXiv preprint arXiv:2307.04577},
  year={2023}
}

@inproceedings{carfi2024modular,
  title={A modular architecture for IMU-based data gloves},
  author={Carf{\`\i}, Alessandro and Alameh, Mohamad and Belcamino, Valerio and Mastrogiovanni, Fulvio},
  booktitle={European Robotics Forum},
  pages={53--57},
  year={2024},
  organization={Springer}
}

@inproceedings{grauman2024ego,
  title={Ego-exo4d: Understanding skilled human activity from first-and third-person perspectives},
  author={Grauman, Kristen and Westbury, Andrew and Torresani, Lorenzo and Kitani, Kris and Malik, Jitendra and Afouras, Triantafyllos and Ashutosh, Kumar and Baiyya, Vijay and Bansal, Siddhant and Boote, Bikram and others},
  booktitle={Proceedings of the IEEE/CVF Conference on Computer Vision and Pattern Recognition},
  pages={19383--19400},
  year={2024}
}

@misc{jain2024vid2robotendtoendvideoconditionedpolicy,
      title={Vid2Robot: End-to-end Video-conditioned Policy Learning with Cross-Attention Transformers}, 
      author={Vidhi Jain and Maria Attarian and Nikhil J Joshi and Ayzaan Wahid and Danny Driess and Quan Vuong and Pannag R Sanketi and Pierre Sermanet and Stefan Welker and Christine Chan and Igor Gilitschenski and Yonatan Bisk and Debidatta Dwibedi},
      year={2024},
      eprint={2403.12943},
      archivePrefix={arXiv},
      primaryClass={cs.RO},
      url={https://arxiv.org/abs/2403.12943}, 
}

@article{zhou2025you,
  title={You Only Teach Once: Learn One-Shot Bimanual Robotic Manipulation from Video Demonstrations},
  author={Zhou, Huayi and Wang, Ruixiang and Tai, Yunxin and Deng, Yueci and Liu, Guiliang and Jia, Kui},
  journal={arXiv preprint arXiv:2501.14208},
  year={2025}
}

@misc{zakka2021xirlcrossembodimentinversereinforcement,
      title={XIRL: Cross-embodiment Inverse Reinforcement Learning}, 
      author={Kevin Zakka and Andy Zeng and Pete Florence and Jonathan Tompson and Jeannette Bohg and Debidatta Dwibedi},
      year={2021},
      eprint={2106.03911},
      archivePrefix={arXiv},
      primaryClass={cs.RO},
      url={https://arxiv.org/abs/2106.03911}, 
}

@misc{wang2023mimicplaylonghorizonimitationlearning,
      title={MimicPlay: Long-Horizon Imitation Learning by Watching Human Play}, 
      author={Chen Wang and Linxi Fan and Jiankai Sun and Ruohan Zhang and Li Fei-Fei and Danfei Xu and Yuke Zhu and Anima Anandkumar},
      year={2023},
      eprint={2302.12422},
      archivePrefix={arXiv},
      primaryClass={cs.RO},
      url={https://arxiv.org/abs/2302.12422}, 
}

@misc{zhu2024visionbasedmanipulationsinglehuman,
      title={Vision-based Manipulation from Single Human Video with Open-World Object Graphs}, 
      author={Yifeng Zhu and Arisrei Lim and Peter Stone and Yuke Zhu},
      year={2024},
      eprint={2405.20321},
      archivePrefix={arXiv},
      primaryClass={cs.RO},
      url={https://arxiv.org/abs/2405.20321}, 
}

@misc{brohan2023rt1roboticstransformerrealworld,
      title={RT-1: Robotics Transformer for Real-World Control at Scale}, 
      author={Anthony Brohan and Noah Brown and Justice Carbajal and Yevgen Chebotar and Joseph Dabis and Chelsea Finn and Keerthana Gopalakrishnan and Karol Hausman and Alex Herzog and Jasmine Hsu and Julian Ibarz and Brian Ichter and Alex Irpan and Tomas Jackson and Sally Jesmonth and Nikhil J Joshi and Ryan Julian and Dmitry Kalashnikov and Yuheng Kuang and Isabel Leal and Kuang-Huei Lee and Sergey Levine and Yao Lu and Utsav Malla and Deeksha Manjunath and Igor Mordatch and Ofir Nachum and Carolina Parada and Jodilyn Peralta and Emily Perez and Karl Pertsch and Jornell Quiambao and Kanishka Rao and Michael Ryoo and Grecia Salazar and Pannag Sanketi and Kevin Sayed and Jaspiar Singh and Sumedh Sontakke and Austin Stone and Clayton Tan and Huong Tran and Vincent Vanhoucke and Steve Vega and Quan Vuong and Fei Xia and Ted Xiao and Peng Xu and Sichun Xu and Tianhe Yu and Brianna Zitkovich},
      year={2023},
      eprint={2212.06817},
      archivePrefix={arXiv},
      primaryClass={cs.RO},
      url={https://arxiv.org/abs/2212.06817}, 
}

@inproceedings{zitkovich2023rt,
  title={Rt-2: Vision-language-action models transfer web knowledge to robotic control},
  author={Zitkovich, Brianna and Yu, Tianhe and Xu, Sichun and Xu, Peng and Xiao, Ted and Xia, Fei and Wu, Jialin and Wohlhart, Paul and Welker, Stefan and Wahid, Ayzaan and others},
  booktitle={Conference on Robot Learning},
  pages={2165--2183},
  year={2023},
  organization={PMLR}
}

@article{kim2024openvla,
  title={Openvla: An open-source vision-language-action model},
  author={Kim, Moo Jin and Pertsch, Karl and Karamcheti, Siddharth and Xiao, Ted and Balakrishna, Ashwin and Nair, Suraj and Rafailov, Rafael and Foster, Ethan and Lam, Grace and Sanketi, Pannag and others},
  journal={arXiv preprint arXiv:2406.09246},
  year={2024}
}

@misc{octomodelteam2024octoopensourcegeneralistrobot,
      title={Octo: An Open-Source Generalist Robot Policy}, 
      author={Octo Model Team and Dibya Ghosh and Homer Walke and Karl Pertsch and Kevin Black and Oier Mees and Sudeep Dasari and Joey Hejna and Tobias Kreiman and Charles Xu and Jianlan Luo and You Liang Tan and Lawrence Yunliang Chen and Pannag Sanketi and Quan Vuong and Ted Xiao and Dorsa Sadigh and Chelsea Finn and Sergey Levine},
      year={2024},
      eprint={2405.12213},
      archivePrefix={arXiv},
      primaryClass={cs.RO},
      url={https://arxiv.org/abs/2405.12213}, 
}

@misc{black2024pi0visionlanguageactionflowmodel,
      title={$\pi_0$: A Vision-Language-Action Flow Model for General Robot Control}, 
      author={Kevin Black and Noah Brown and Danny Driess and Adnan Esmail and Michael Equi and Chelsea Finn and Niccolo Fusai and Lachy Groom and Karol Hausman and Brian Ichter and Szymon Jakubczak and Tim Jones and Liyiming Ke and Sergey Levine and Adrian Li-Bell and Mohith Mothukuri and Suraj Nair and Karl Pertsch and Lucy Xiaoyang Shi and James Tanner and Quan Vuong and Anna Walling and Haohuan Wang and Ury Zhilinsky},
      year={2024},
      eprint={2410.24164},
      archivePrefix={arXiv},
      primaryClass={cs.LG},
      url={https://arxiv.org/abs/2410.24164}, 
}

@misc{intelligence2025pi05visionlanguageactionmodelopenworld,
      title={$\pi_{0.5}$: a Vision-Language-Action Model with Open-World Generalization}, 
      author={Physical Intelligence and Kevin Black and Noah Brown and James Darpinian and Karan Dhabalia and Danny Driess and Adnan Esmail and Michael Equi and Chelsea Finn and Niccolo Fusai and Manuel Y. Galliker and Dibya Ghosh and Lachy Groom and Karol Hausman and Brian Ichter and Szymon Jakubczak and Tim Jones and Liyiming Ke and Devin LeBlanc and Sergey Levine and Adrian Li-Bell and Mohith Mothukuri and Suraj Nair and Karl Pertsch and Allen Z. Ren and Lucy Xiaoyang Shi and Laura Smith and Jost Tobias Springenberg and Kyle Stachowicz and James Tanner and Quan Vuong and Homer Walke and Anna Walling and Haohuan Wang and Lili Yu and Ury Zhilinsky},
      year={2025},
      eprint={2504.16054},
      archivePrefix={arXiv},
      primaryClass={cs.LG},
      url={https://arxiv.org/abs/2504.16054}, 
}

@misc{chi2024universalmanipulationinterfaceinthewild,
      title={Universal Manipulation Interface: In-The-Wild Robot Teaching Without In-The-Wild Robots}, 
      author={Cheng Chi and Zhenjia Xu and Chuer Pan and Eric Cousineau and Benjamin Burchfiel and Siyuan Feng and Russ Tedrake and Shuran Song},
      year={2024},
      eprint={2402.10329},
      archivePrefix={arXiv},
      primaryClass={cs.RO},
      url={https://arxiv.org/abs/2402.10329}, 
}

@article{qiu2025humanoid,
  title={Humanoid Policy\~{} Human Policy},
  author={Qiu, Ri-Zhao and Yang, Shiqi and Cheng, Xuxin and Chawla, Chaitanya and Li, Jialong and He, Tairan and Yan, Ge and Yoon, David J and Hoque, Ryan and Paulsen, Lars and others},
  journal={arXiv preprint arXiv:2503.13441},
  year={2025}
}

@article{niu2025human2locoman,
  title={Human2locoman: Learning versatile quadrupedal manipulation with human pretraining},
  author={Niu, Yaru and Zhang, Yunzhe and Yu, Mingyang and Lin, Changyi and Li, Chenhao and Wang, Yikai and Yang, Yuxiang and Yu, Wenhao and Zhang, Tingnan and Li, Zhenzhen and others},
  journal={arXiv preprint arXiv:2506.16475},
  year={2025}
}

@article{li2025h2r,
  title={H2R: A Human-to-Robot Data Augmentation for Robot Pre-training from Videos},
  author={Li, Guangrun and Lyu, Yaoxu and Liu, Zhuoyang and Hou, Chengkai and Zhang, Jieyu and Zhang, Shanghang},
  journal={arXiv preprint arXiv:2505.11920},
  year={2025}
}

@article{yang2025egovla,
  title={Egovla: Learning vision-language-action models from egocentric human videos},
  author={Yang, Ruihan and Yu, Qinxi and Wu, Yecheng and Yan, Rui and Li, Borui and Cheng, An-Chieh and Zou, Xueyan and Fang, Yunhao and Cheng, Xuxin and Qiu, Ri-Zhao and others},
  journal={arXiv preprint arXiv:2507.12440},
  year={2025}
}

@article{liu2025egozero,
  title={Egozero: Robot learning from smart glasses},
  author={Liu, Vincent and Adeniji, Ademi and Zhan, Haotian and Haldar, Siddhant and Bhirangi, Raunaq and Abbeel, Pieter and Pinto, Lerrel},
  journal={arXiv preprint arXiv:2505.20290},
  year={2025}
}

@inproceedings{o2024open,
  title={Open x-embodiment: Robotic learning datasets and rt-x models: Open x-embodiment collaboration 0},
  author={O’Neill, Abby and Rehman, Abdul and Maddukuri, Abhiram and Gupta, Abhishek and Padalkar, Abhishek and Lee, Abraham and Pooley, Acorn and Gupta, Agrim and Mandlekar, Ajay and Jain, Ajinkya and others},
  booktitle={2024 IEEE International Conference on Robotics and Automation (ICRA)},
  pages={6892--6903},
  year={2024},
  organization={IEEE}
}

@article{zhao2023learning,
  title={Learning fine-grained bimanual manipulation with low-cost hardware},
  author={Zhao, Tony Z and Kumar, Vikash and Levine, Sergey and Finn, Chelsea},
  journal={arXiv preprint arXiv:2304.13705},
  year={2023}
}

@inproceedings{qin2022dexmv,
  title={Dexmv: Imitation learning for dexterous manipulation from human videos},
  author={Qin, Yuzhe and Wu, Yueh-Hua and Liu, Shaowei and Jiang, Hanwen and Yang, Ruihan and Fu, Yang and Wang, Xiaolong},
  booktitle={European Conference on Computer Vision},
  pages={570--587},
  year={2022},
  organization={Springer}
}

@inproceedings{pavlakos2024reconstructing,
  title={Reconstructing hands in 3d with transformers},
  author={Pavlakos, Georgios and Shan, Dandan and Radosavovic, Ilija and Kanazawa, Angjoo and Fouhey, David and Malik, Jitendra},
  booktitle={Proceedings of the IEEE/CVF Conference on Computer Vision and Pattern Recognition},
  pages={9826--9836},
  year={2024}
}

@misc{peebles2023scalablediffusionmodelstransformers,
      title={Scalable Diffusion Models with Transformers}, 
      author={William Peebles and Saining Xie},
      year={2023},
      eprint={2212.09748},
      archivePrefix={arXiv},
      primaryClass={cs.CV},
      url={https://arxiv.org/abs/2212.09748}, 
}

@article{nair2022r3m,
  title={R3m: A universal visual representation for robot manipulation},
  author={Nair, Suraj and Rajeswaran, Aravind and Kumar, Vikash and Finn, Chelsea and Gupta, Abhinav},
  journal={arXiv preprint arXiv:2203.12601},
  year={2022}
}

@inproceedings{christen2024synh2r,
  title={Synh2r: Synthesizing hand-object motions for learning human-to-robot handovers},
  author={Christen, Sammy and Feng, Lan and Yang, Wei and Chao, Yu-Wei and Hilliges, Otmar and Song, Jie},
  booktitle={2024 IEEE International Conference on Robotics and Automation (ICRA)},
  pages={3168--3175},
  year={2024},
  organization={IEEE}
}

@inproceedings{bharadhwaj2024roboagent,
  title={Roboagent: Generalization and efficiency in robot manipulation via semantic augmentations and action chunking},
  author={Bharadhwaj, Homanga and Vakil, Jay and Sharma, Mohit and Gupta, Abhinav and Tulsiani, Shubham and Kumar, Vikash},
  booktitle={2024 IEEE International Conference on Robotics and Automation (ICRA)},
  pages={4788--4795},
  year={2024},
  organization={IEEE}
}
